\pdfoutput=1

\documentclass[11pt]{article}

\usepackage[]{emnlp2021}

\usepackage{times}
\usepackage{latexsym}

\usepackage[T1]{fontenc}

\usepackage[utf8]{inputenc}
\usepackage{booktabs}       
\usepackage{url}            
\usepackage{amsfonts}       
\usepackage{nicefrac}       
\usepackage{amsmath}
\usepackage{graphicx}
\usepackage{bm, upgreek}
\usepackage{amssymb}
\usepackage{array, booktabs}
\usepackage{multirow}
\usepackage{mathtools}
\usepackage{arydshln}
\usepackage{tabularx}
\usepackage{adjustbox}
\usepackage{color}
\usepackage{soul}
\usepackage{caption}
\usepackage{xcolor}
\usepackage{pifont}
\usepackage{comment}
\usepackage{cleveref}
\usepackage[ruled,vlined]{algorithm2e}
\usepackage{makecell}
\usepackage{diagbox}
\usepackage{fnpct}
\usepackage{cleveref}

\usepackage{microtype}
\newcommand{\Ours}{\textbf{Bio}medical \textbf{LA}nguage \textbf{M}odel \textbf{A}nalysis}
\newcommand{\ours}{\textsc{BioLAMA}}

\crefformat{section}{\S#2#1#3} 
\crefformat{subsection}{\S#2#1#3}
\crefformat{subsubsection}{\S#2#1#3}

\newcommand\blfootnote[1]{%
  \begingroup
  \renewcommand\thefootnote{}\footnote{#1}%
  \addtocounter{footnote}{-1}%
  \endgroup
}

\graphicspath{ {./figures/} }

\title{Can Language Models be Biomedical Knowledge Bases?}

\author{
  Mujeen Sung$^{1}$\quad Jinhyuk Lee$^{1,2 \dagger}$\quad Sean S. Yi$^{1}$ \\ \textbf{\quad \quad Minji Jeon$^{3}$\quad Sungdong Kim$^{4}$\quad Jaewoo Kang$^{1 \dagger}$\quad} \\
  Korea University$^{1}$\quad Princeton University$^{2}$ \\\quad Icahn School of Medicine at Mount Sinai$^{3}$\quad NAVER AI Lab$^{4}$\\
  \texttt{\{mujeensung,jinhyuk\_lee,seanswyi,kangj\}@korea.ac.kr} \\
  \texttt{minji.jeon@mssm.edu} \\
  \texttt{sungdong.kim@navercorp.com}}

\begin{document}
\maketitle

\begin{abstract}
Pre-trained language models (LMs) have become ubiquitous in solving various natural language processing (NLP) tasks.
There has been increasing interest in what knowledge these LMs contain and how we can extract that knowledge, treating LMs as knowledge bases (KBs).
While there has been much work on probing LMs in the general domain, there has been little attention to whether these powerful LMs can be used as domain-specific KBs.
To this end, we create the \ours~benchmark, which is comprised of 49K biomedical factual knowledge triples for probing biomedical LMs.
We find that biomedical LMs with recently proposed probing methods can achieve up to 18.51\% Acc@5 on retrieving biomedical knowledge.
Although this seems promising given the task difficulty, our detailed analyses reveal that most predictions are highly correlated with prompt templates without any subjects, hence producing similar results on each relation and hindering their capabilities to be used as domain-specific KBs.
We hope that \ours~can serve as a challenging benchmark for biomedical factual probing.\footnote{\href{https://github.com/dmis-lab/BioLAMA}{https://github.com/dmis-lab/BioLAMA}}
\end{abstract}
\blfootnote{\textsuperscript{$\dagger$}Corresponding authors.}
\section{Introduction}

Recent success in natural language processing can be largely attributed to powerful pre-trained language models (LMs) that learn contextualized representations of words from large amounts of unstructured corpora~\citep{peters2018deep,devlin2019bert}.
There have been recent works in probing how much knowledge these LMs contain in their parameters~\citep{petroni2019language} and how to effectively extract such knowledge.~\citep{shin2020autoprompt,jiang2020can,zhong2021factual}.

\begin{figure}
    \centering
    \includegraphics[scale=0.60]{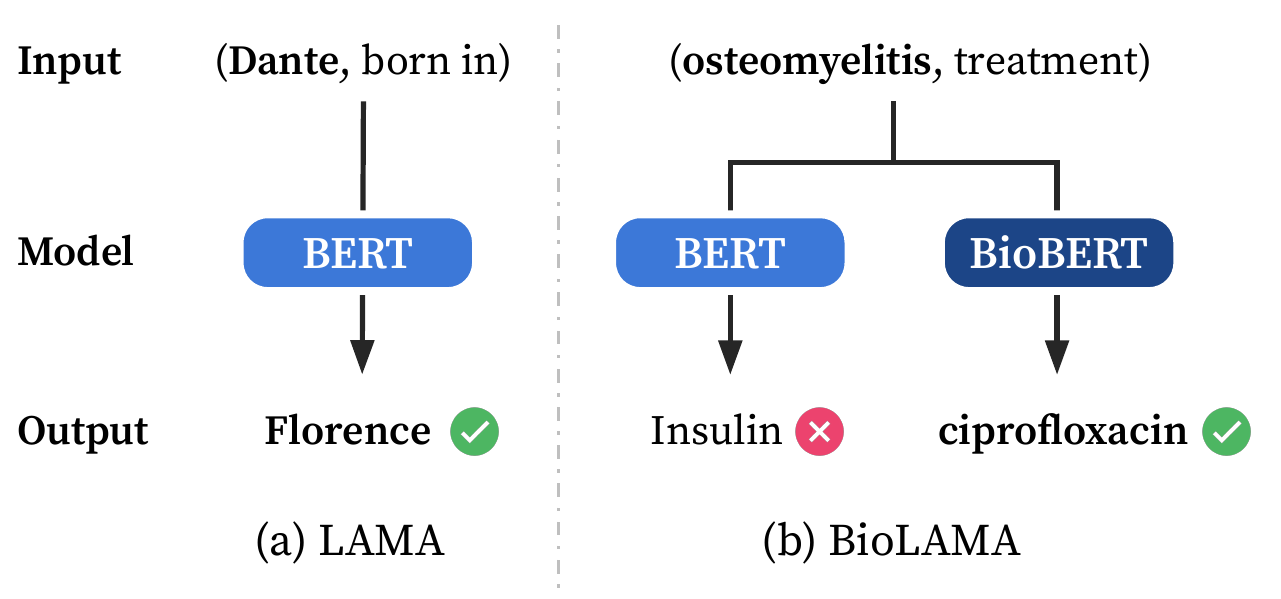}
    \caption{
        Comparison of \textsc{LAMA}~\citep{petroni2019language} and \ours.
        (a)~\textsc{LAMA} tests general knowledge of LMs. (b)~\ours~tests expert-level biomedical knowledge of LMs such as a treatment for a disease.
    }
    \vspace{-0.3cm}
    \label{fig:biolama_overview}
\end{figure}
\begin{table*}[t]
    \centering
    \resizebox{1.95\columnwidth}{!}{%
    \begin{tabular}{llll}
        \toprule
        
        & \textbf{Relation Name} & \textbf{Manual Prompt} & \textbf{Object Answer} \\
        
        \midrule
        
        \multirow{4}{*}{{\textsc{LAMA}}} & \multicolumn{3}{l}{\textbf{\# Relations}: 41 \quad \textbf{ \# Entity Types}: 25$^*$ \quad \textbf{\# Triples}: 41k \quad \textbf{Sources}: Wikidata} \\
        
        \cmidrule{2-4}
        
        & place of birth & \underline{Dante} was born in [Y]. & Florence \\ 
        
        & place of death & \underline{Adolphe Adam} died in [Y]. & Paris \\
        
        & official language & The official language of \underline{Mauritius} is [Y]. & English \\
        
        \midrule
        
        \multirow{4}{*}{{\ours}} & \multicolumn{3}{l}{\textbf{\# Relations}: 36 \quad \textbf{ \# Entity Types}: 12 \quad \textbf{\# Triples}: 49K \quad \textbf{Sources}: CTD, UMLS, Wikidata} \\
        
        \cmidrule{2-4}
        
        & medical condition treated & \underline{Amantadine} has effects on [Y]. & Parkinson's disease, ... \\
        
        & symptoms & \underline{Hepatitis} has symptoms such as [Y]. & abdominal pain, ... \\
        
        & affects binding & \underline{Nicotine} binds to [Y]. & CHRNA4, CHRNB2, ...\\
        
        \bottomrule
    \end{tabular}
    }
    \caption{Comparison of \textsc{LAMA} (T-REx) and \ours~with their statistics. For each dataset, we also show the examples of relations and their corresponding manual prompts and answers. The underlined entities are subjects and [Y] refers to the object to be predicted. 
    $^*$: obtained from \citet{cao2021knowledgeable}.
    }
    \vspace{-0.3cm}
    \label{tab:compare_lama_biolama}
\end{table*}

While factual probing of LMs has attracted much attention from researchers, a more practical application would be to leverage the power of domain-specific LMs~\citep{beltagy2019scibert, lee2020biobert} as domain knowledge bases (KBs).
Unlike recent works that probe general domain knowledge, we ask whether it is also possible to retrieve expert knowledge from LMs.
Specifically, we tune our focus on factual knowledge probing for the biomedical domain as shown in \Cref{fig:biolama_overview}.

To inspect the potential utility of LMs as biomedical KBs, we create and release the \Ours~(\ours) probe. \ours~consists of 49K biomedical factual triples whose relations have been manually curated from three different knowledge sources: the Comparative Toxicogenomics Database (CTD), the Unified Medical Language System (UMLS), and Wikidata.
While our biomedical factual triples are inherently more difficult to probe (see \Cref{tab:compare_lama_biolama} for examples), \ours~also poses technical challenges such as multi-token object decoding.

Initial probing results on \ours~show that the best performing LM achieves up to 7.28\% Acc@1 and 18.51\% Acc@5, and outperforms an information extraction (IE) baseline~\citep{lee2016best}.
Although this result seems promising, we find that their output distributions are largely biased to a small number of entities in each relation.
Along this line, we use two metrics, prompt bias~\citep{cao2021knowledgeable} and synonym variance, to investigate the behavior of LMs as KBs.
Our analysis shows that while LMs seem to be more aware of synonyms than the IE baseline, they output highly biased predictions given the prompt template of each relation.
Our result calls for better LMs and probing methods that can retrieve rich but still useful biomedical entities.

\section {\ours}
\label{sec:biolama}

In this section, we detail the construction of \ours~including the data curation process and pre-processing steps.
Statistics and examples of \ours~are shown in \Cref{tab:compare_lama_biolama} along with those from \textsc{LAMA}~\citep{petroni2019language}.

\subsection{Knowledge Sources}

\paragraph{CTD}
The CTD\footnote{\href{http://ctdbase.org/}{http://ctdbase.org/}} is a public biomedical database on relationships and interactions between biomedical entities such as diseases, chemicals, and genes~\citep{davis2020ctd}.
It provides both manually curated and automatically inferred triples in English, and we only use the manually curated triples for a better quality of our dataset.
We use the April 1st, 2021 version of the CTD.

\paragraph{UMLS}
The UMLS Metathesaurus\footnote{\href{https://www.nlm.nih.gov/research/umls/}{https://www.nlm.nih.gov/research/umls/}} is a large-scale database that provides information regarding various concepts and vocabularies in the biomedical domain~\citep{bodenreider2004unified}.
We use the 2020AB version of the UMLS.
The UMLS provides entity names in various languages and we use the ones in English.

\paragraph{Wikidata}
Wikidata\footnote{\href{https://wikidata.org}{https://wikidata.org}} is a public KB with items across various domains.
Following the previous works~\citep{turki2019wikidata,waagmeester2020science}, we retrieve biomedical entities and relations using SPARQL queries.
We use the dump of the January 25th, 2021 version.
Similar to the UMLS, we use entity names in English.

\begin{table}[t]
    \centering
    \resizebox{0.95\columnwidth}{!}{%
    \begin{tabular}{lcc}
        \toprule
\textbf{Dataset} & \textbf{Obj in Sbj (\%)} & \textbf{\# Object Subwords} \\
\midrule
{\textsc{LAMA}} & 12.81 & 1.00 \\
{\textsc{LAMA-UHN}} & \textbf{0.00} & 1.00 \\
{\textsc{X-FACTR}} & 6.35 & 3.07 \\
{\ours} & \textbf{0.00} & \textbf{4.52} \\
 \bottomrule
    \end{tabular}
    }
    \caption{Comparison of probing benchmarks: ratio of subjects with objects as substrings, and the average subword numbers of object entities. We compare these two aspects of~\ours~to \textsc{LAMA}, \textsc{LAMA-UHN}~\citep{poerner2020bert} and \textsc{X-FACTR}~\citep{jiang2020x}.
    }
    \vspace{-0.3cm}
    \label{tab:dataset_comparision}
\end{table}
\begin{table*}[t]
    \centering
    \resizebox{2.0\columnwidth}{!}{%
    \begin{tabular}{lccccccc}
        \toprule
        
        \multirow{2}{*}{\textbf{Source}} & 
        \multirow{2}{*}{\textbf{IE}} & \multicolumn{2}{c}{\textbf{BERT}} & \multicolumn{2}{c}{\textbf{BioBERT}} & \multicolumn{2}{c}{\textbf{Bio-LM}} \\
        
        & & Manual & Opti. & Manual & Opti. & Maual & Opti. \\
        \midrule
        
        \textbf{CTD} (11.13\%)
        & \textbf{5.06} / \textbf{12.15} 
        & 0.06 / 1.20 
        & 3.56 / 6.97 
        & 0.42 / 3.25 
        & \underline{4.82} / 9.74 
        & 1.77 / 7.30 
        & 2.99 / \underline{10.19} 
        \\
        
        \textbf{UMLS} (9.67\%)
        & 3.53 / 6.99 
        & 0.82 / 1.99 
        & 1.44 / 3.65 
        & 1.16 / 3.82 
        & \underline{5.08} / \underline{13.28} 
        & 3.44 / 8.88 
        & \textbf{8.25} / \textbf{20.19} 
        \\
        
        \textbf{Wikidata} (5.76\%)
        & 7.03 / 15.55 
        & 1.16 / 6.04 
        & 3.29 / 8.13 
        & 3.67 / 11.20 
        & 4.21 / 12.91 
        & \textbf{11.97} / \textbf{25.92} 
        & \underline{10.60} / \underline{25.15} 
        \\
        
        \midrule
        
        \textbf{Average}
        & 5.21 / 11.56 
        & 0.86 / 3.08 
        & 2.76 / 6.25 
        & 1.75 / 6.09 
        & 4.70 / 11.98 
        & \underline{5.72} / \underline{14.03} 
        & \textbf{7.28} / \textbf{18.51} 
        \\
        
        \bottomrule
    \end{tabular}
    }
    \caption{Main experimental results on~\ours. 
    We report Acc@1/Acc@5 of each model including the macro average across three different knowledge sources.
    We also report ratios of the majority objects in each knowledge source (averaged over its relations) in the parentheses.
    Highest and second-highest scores are \textbf{boldfaced} and \underline{underlined}, respectively.
    Manual: manual prompt. Opti.: OptiPrompt.
    The results of OptiPrompt are the mean of 5 runs with different seeds.
    See \Cref{sec:all_results} for the performance on each relation.
    }
    \vspace{-0.3cm}
    \label{tab:probing_method_experiments}
\end{table*}

\subsection{Data Pre-processing}\label{sec:main_prepro}

From our initial factual triples from the knowledge sources above, we apply several pre-processing steps to further improve the quality of~\ours.
First, considering the trade-off between the coverage and difficulty of probing, we restrict the lengths of entities to be \(\le\)10 subwords, which covers 90\% of the entities.\footnote{Based on the \texttt{BERT-base-cased} tokenizer.}
Note that \textsc{LAMA} only contains single-token objects, which makes the task easier, but less practical.
Following~\citet{poerner2020bert}, we also discard easy triples where objects are substrings of the paired subjects~(e.g., ``iron deficiency''-``iron''), which prevents trivial solutions using the surface forms of the subjects.
For each relation, we split samples into training, development, and test sets with a 40:10:50 ratio.
The training set is provided for learning or finding good prompts for each relation.
More details on pre-processing steps are available in \Cref{sec:apdx_prepro}.

After the pre-processing, we are able to obtain 22K triples with 15 relations from the CTD, 21.2K triples with 16 relations from the UMLS, and 5.8K triples with 5 relations from Wikidata~(see \Cref{sec:statistics_of_biolama} for the detailed statistics).
In \Cref{tab:dataset_comparision}, we compare various probing benchmarks with \ours.
By design, \ours~has no objects that are substrings of their subjects and object entities are much longer on average, which makes our benchmark challenging but much more practical.

\paragraph{Evaluation Metric}
We use top-$k$ accuracy (Acc@$k$), which is 1 if any of the top $k$ object entities are included in the annotated object list, and is 0 otherwise.
We use both Acc@1 and Acc@5 since most biomedical entities are related to multiple biomedical entities~(i.e., \textit{N-to-M} relations).

\section{Experiment}
\label{sec:experiments}
\begin{table*}[t]
    \centering
    \resizebox{2.0\columnwidth}{!}{%
\begin{tabular}{lll}
\toprule
\textbf{Relation ID} & \textbf{Subject} & \textbf{Top 5 Predictions} \\
\midrule
\multirow{3}{*}{\begin{tabular}[c]{l}UMLS - UR254 \\ (27.71 / 38.41)\end{tabular}} & \multicolumn{2}{l}{[X] has symptoms such as [Y].} \\
\cmidrule{2-3}
& Pituicytoma & \textbf{headache}, headaches, pain, bone pain, pain and bleeding \\
& Intravascular fasciitis & \textbf{pain}, pain and swelling, swelling and pain, swelling, edema \\
& Microfollicular adenoma & headache, epistaxis, pruritus, itching, flushing \\
& Parosteal Osteosarcoma & pain, bone pain, pain and swelling, swelling and pain, pain and bleeding \\
  \midrule
\multirow{3}{*}{\begin{tabular}[c]{l}CTD - CG4\\ (8.42 / 20.59)\end{tabular}} & \multicolumn{2}{l}{[X] results in increased activity of [Y] protein.} \\
\cmidrule{2-3}
& Dieldrin & \textbf{ESR1}, \textbf{NR1I2}, NR3C1, CASP1, PPARÎ³ \\
& isofenphos & \textbf{ESR1}, NR1I2, PPARÎ³, CYP1A2, CDKN1A1 \\
& Dithiothreitol & ESR1, NR1I2, NR3C1, NR1I1, CASP1 \\
& Indigo Carmine & ESR1, NR1I2, CYP1A2, NR3C1, CASP1 \\
 \midrule
\multirow{3}{*}{\begin{tabular}[c]{l}Wikidata - P2176\\ (20.14 / 39.57)\end{tabular}} & \multicolumn{2}{l}{The standard treatment for patients with [X] is a drug such as [Y].} \\
\cmidrule{2-3}
& Haverhill fever & \textbf{doxycycline}, ciprofloxacin, penicillin, erythromycin, azithromycin \\
& influenza & \textbf{zanamivir}, interferon, \textbf{peramivir}, oseltamivir or peramivir, doxycycline \\
& cryptosporidiosis & amphotericin B, praziquantel, itraconazole, albendazole, fluconazole \\
& tremor & pilocarpine, baclofen, botulinum toxin, diazepam, clonazepam \\
\bottomrule
\end{tabular}
    }
    \caption{Top 5 predictions of Bio-LM (w/ OptiPrompt) given each prompt and different subjects. For each relation, we also report its Acc@1/Acc@5. Correct predictions are in boldface.
    For more examples, see \Cref{sec:predictions_appendix}.
    }\vspace{-0.3cm}\label{tab:predictions}
\end{table*}

\subsection{Models}

\paragraph{Information Extraction}
Many biomedical NLP tools rely on automated IE systems that can provide relevant entities or articles given a query.
In this work, we use the Biomedical Entity Search Tool (BEST)~\citep{lee2016best}\footnote{\href{https://best.korea.ac.kr/}{https://best.korea.ac.kr/}} as an IE system and compare it with LM-based probing methods.
BEST incorporates biomedical entities when building their search index over PubMed, a large-scale biomedical corpus, and returns biomedical entities given a keyword-based query.
To fully make use of BEST, we create AND queries using a subject entity and a lemmatized relation name (e.g., ``(meclozine) AND (medical condition treat)''), and use retrieved entities as its predictions.

\paragraph{Language Models}
We use one general-domain LM and two biomedical LMs: BERT~\citep{devlin2019bert}, BioBERT~\citep{lee2020biobert}\footnote{Since existing checkpoints of BioBERT do not contain LM heads for probing, we pre-train another BioBERT (\texttt{biobert-base-cased-v1.2}), which is the same as the previous version of BioBERT but with an LM head.}, and Bio-LM~\citep{lewis2020pretrained}\footnote{\texttt{RoBERTa-base-PM-Voc}}.
BioBERT and Bio-LM are both pre-trained over PubMed.
While Bio-LM also uses a custom vocabulary learned from PubMed, BioBERT uses the same vocabulary as BERT, which enables the continual learning of BioBERT initialized from BERT.

\subsection{Probing Methods}

\paragraph{Prompts}
We use a fill-in-the-blank cloze statement (i.e., a  ``prompt'') for probing and choose two different methods of prompt generation: manual prompts~\citep{petroni2019language} and OptiPrompt~\citep{zhong2021factual}.
For each relation, we first create manual prompts with domain experts~(\Cref{sec:manual_prompts}).
On the other hand, OptiPrompt automatically learns continuous embeddings that can better extract factual knowledge for each relation, which are trained with our training examples.
Following~\citet{zhong2021factual}, we initialize the continuous embeddings with the embeddings of manual prompts, which worked consistently better than random initialization in our experiments.

\paragraph{Multi-token Object Decoding}

Since the majority of entities in \ours~are made up of multiple tokens, we implement a multi-token decoding strategy following~\citet{jiang2020x}.
Among their decoding methods, we use the confidence-based method which produced the best results.
The confidence-based method greedily decodes output tokens sorted by the maximum logit in each token position.
Note that we do not restrict our output spaces by any pre-defined sets of biomedical entities since we are more interested in how accurately the LMs contain biomedical knowledge in an unconstrained setting.\footnote{Using a pre-defined set of object entities removes the necessity of using complicated decoding strategies and will possibly improve the probing accuracy as well, which we leave as future work.}
See \Cref{sec:implementation_details} for the implementation details of our decoding method.

\subsection{Main Results}

Experimental results on \ours~are summarized in \Cref{tab:probing_method_experiments}.
First, BioBERT and Bio-LM are both able to retrieve factual information better than BERT, which demonstrates the effectiveness of domain-specific pre-training.
Also, Bio-LM shows consistently better performance than BioBERT (BERT < BioBERT < Bio-LM).
We believe that this may be attributed to the custom vocabulary of Bio-LM learned from a biomedical corpus.
Using OptiPrompt also shows consistent improvement over manual prompts in all LMs.
Notably, the IE system is able to achieve the best performance on the CTD relations, but performs worse than BioBERT and Bio-LM on the UMLS and Wikidata relations.

While we are able to achieve 18.51\% Acc@5 with Bio-LM (w/ OptiPrompt) on average, note that the average Acc@1s on the CTD and UMLS relations are lower than majority voting (e.g., 9.67\% (majority) vs. 8.25\% Acc@1 (Bio-LM) in UMLS), which shows the difficulty of accurately extracting biomedical facts from these models.
\vspace{-0.05cm}
\section{LMs are Not Biomedical KBs, Yet} \label{sec:analysis}

In this section, we thoroughly inspect the predictions of Bio-LM (w/ OptiPrompt) and quantitatively characterize the behavior of each model.
Our analyses suggest that we might need stronger biomedical LMs and probing methods to make use of these LMs as domain-specific knowledge bases.

\subsection{Predictions}

In \Cref{tab:predictions}, we present two correct and two incorrect predictions for three different relations where Bio-LM (w/ OptiPrompt) achieves high accuracy.
One aspect that stands out is that predictions tend to be highly biased towards a few objects (e.g., ``headache'', ``pain'', or ``ESR1'').
Motivated by this observation, we further measure two metrics that can characterize the behavior of each model in detail: prompt bias and synonym variance.

\subsection{How Biomedical LMs Predict}\label{sec:metrics}

\begin{figure}[tp]
    \centering
    \includegraphics[width=1\columnwidth]{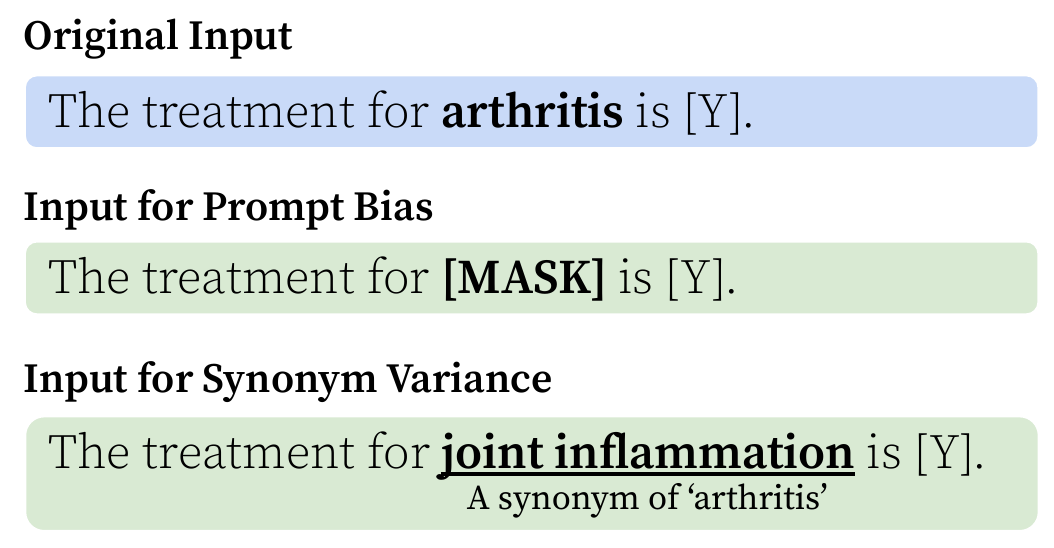}
    \caption{
        Examples of inputs for measuring prompt bias and synonym variance. 
        We use a \texttt{[MASK]} token for the subject when measuring prompt bias, and replace each subject into their synonyms when measuring synonym variance.
    }
    \vspace{-0.3cm}
    \label{fig:prmoptbias_synonymvariance}
\end{figure}

\paragraph{Prompt Bias}
To serve as accurate KBs, LMs must make appropriate object entity predictions given the input subject entity.
\citet{cao2021knowledgeable} quantified prompt biases by measuring how insensitive LMs are to input subjects.
For each relation, we first obtain the probability histogram of each unique object entity being a top-1 prediction \textit{when the subject is given}.
For example, if one relation has 100 test samples and ``pain'' appears 20 times as its top-1 prediction, the probability mass of ``pain'' becomes 20\%.
At the same time, we calculate the probability distribution over unique object entities \textit{when the subject is masked out} (see \Cref{fig:prmoptbias_synonymvariance}).
For instance, a model might assign 30\% to ``pain'' even when the subject is masked out from the prompt.
Prompt bias is the Pearson's correlation coefficient between these two distributions, which indicates how biased the model is to a prompt.
A lower prompt bias means that a model is giving less biased predictions for each relation (i.e., prompt).

\paragraph{Synonym Variance}
Biomedical entities often have a number of synonyms, which are often leveraged for modeling biomedical entity representations~\citep{sung2020biomedical}.
Hence, it is important that predictions over our factual triples do not change when the input subject is replaced by its synonyms.
To assess this aspect, we propose a metric called synonym variance, which measures how much each prediction changes when the subject is replaced with its synonyms (see \Cref{fig:prmoptbias_synonymvariance}).
We create 10 copies of our datasets by replacing the subjects with one of their synonyms chosen randomly.
Synonym variance is the standard deviation of Acc@5 calculated from these new test sets.
Lower synonym variance means that a model is giving more consistent predictions even with different synonyms.

\begin{figure}[t]
\begin{minipage}[c]{0.45\textwidth}
    \centering
    \includegraphics[width=1.0\columnwidth]{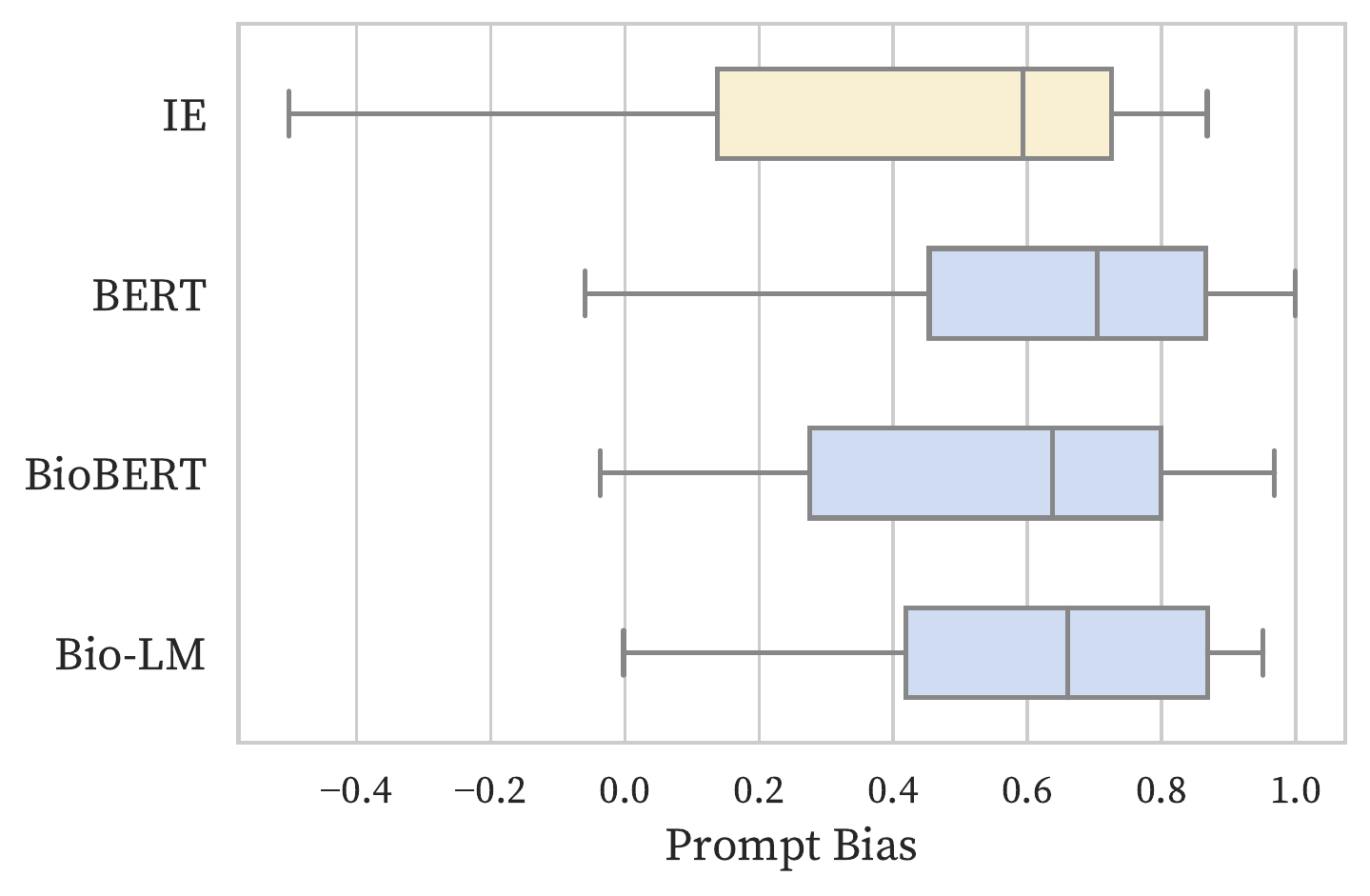}
    \caption{
        Prompt bias of each model. Low prompt bias means that a model is less biased on each prompt.
        See \Cref{sec:metrics} for more details of the metric.
    }
    \vspace{+0.3cm}
    \label{fig:prmoptbias}
\end{minipage}
\\
\begin{minipage}[c]{0.45\textwidth}
    \centering
    \includegraphics[width=1.0\columnwidth]{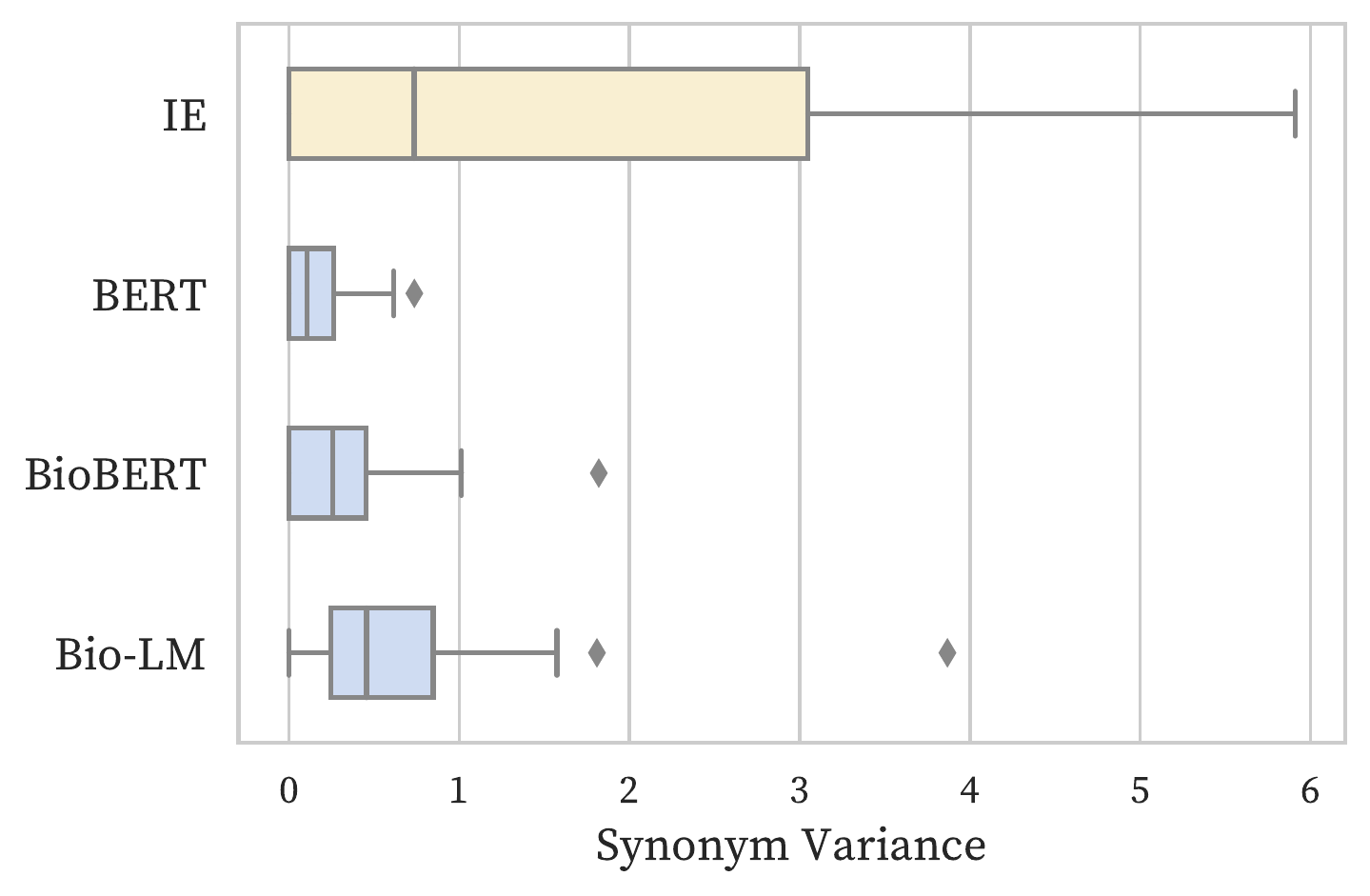}
    \caption{
        Synonym variance of each model. Low synonym variance means that a model gives consistent predictions when the subjects are changed to synonyms.
        See \Cref{sec:metrics} for more details of the metric.
    }
    \vspace{-0.3cm}
    \label{fig:synonymvariance}
\end{minipage}%
\end{figure}

\paragraph{Results}

\Cref{fig:prmoptbias} shows the results of prompt biases in four different models.
Compared to the IE system, the LMs have relatively higher correlations~(over 0.6) meaning that their predictions are more biased towards the prompts.
On the other hand, in \Cref{fig:synonymvariance}, LMs show relatively lower standard deviations over variations of synonyms than the IE system does.
While this can be interpreted that the LMs are more robust to synonym variations, it might also be the result of strong biases in LMs on their prompts.
For example, while BERT has the smallest synonym variance, it has the largest prompt bias, meaning that it is not a synonym-aware model, but just a highly biased model.
\vspace{-0.05cm}
\section{Conclusion}

In this work, we explore the possibility of using LMs as biomedical KBs.
To this end, we release~\ours~as a probing benchmark to measure how much biomedical knowledge can be extracted from LMs.
While biomedical LMs can extract useful facts to some extent, our analysis shows that this is largely due to their predictions being biased towards certain prompts.
In future work, we plan to overcome the underlying challenges in~\ours~and improve the probing accuracy of LMs.

\section*{Acknowledgements}
This work was supported in part by the ICT Creative Consilience program (IITP-2021-0-01819) supervised by the IITP (Institute for Information \& communications Technology Planning \& Evaluation), National Research Foundation of Korea (NRF-2020R1A2C3010638, NRF-2014M3C9A3063541), and Hyundai Motor Chung Mong-Koo Foundation.
We thank the anonymous reviewers for their insightful comments.
\newpage

\section*{Ethical Considerations}

The aim of factual probing is to verify how much knowledge can be retrieved from language models pre-trained using large amoung of corpora.
Due to a lack of data for factual probing in the biomedical domain, we collected data from widely used knowledge sources: the CTD, the UMLS, and Wikidata.
Although these data have undergone inspection by domain experts, biomedical knowledge is continuously growing and therefore we cannot guarantee that this biomedical knowledge is absolute.
Furthermore, without careful inspection, outputs of these LMs should not be considered as a means of drug recommendation or any other medical activity.
We caution future researchers when using \ours~to keep this caveat in mind.


\bibliography{custom}
\bibliographystyle{acl_natbib}

\clearpage
\appendix
\begin{table*}[t]
    \centering
    \resizebox{2.0\columnwidth}{!}{%
    \begin{tabular}{llllccc}
        \toprule
        \textbf{Relation ID} & \textbf{Relation Name} & \textbf{Subject} & \textbf{Object} & \textbf{Train} & \textbf{Dev} & \textbf{Test} \\
        \midrule
        \textbf{CTD}\\
        \midrule
         CD1 & therapeutic & chemical & disease & 756 & 189 & 945 \\
         CD2 & marker/mechanism & chemical & disease & 723 & 181 & 905 \\
         CG1 & decreases expression & chemical & protein & 550 & 137 & 688 \\
         CG17 & increases expression & chemical & mRNA & 740 & 186 & 926 \\
         CG18 & increases expression & chemical & protein & 680 & 170 & 851 \\
         CG2 & decreases activity & chemical & protein & 718 & 179 & 898 \\
         CG21 & increases phosphorylation & chemical & protein & 206 & 51 & 258 \\
         CG4 & increases activity & chemical & protein & 541 & 135 & 677 \\
         CG6 & decreases expression & chemical & mRNA & 648 & 163 & 811 \\
         CG9 & affects binding & chemical & protein & 352 & 89 & 441 \\
         CP1 & decreases & chemical & phenotype & 504 & 127 & 631 \\
         CP2 & increases & chemical & phenotype & 591 & 148 & 739 \\
         CP3 & affects & chemical & phenotype & 360 & 90 & 451 \\
         GD1 & marker/mechanism & gene & disease & 728 & 182 & 911 \\
         GP1 & association & gene & pathway & 704 & 176 & 881 \\
        \midrule
        \textbf{UMLS} \\
        \midrule
         UR116 & clinically associated with & disease & disease & 668 & 167 & 835 \\
         UR124 & may treat & disease & chemical & 463 & 116 & 580 \\
         UR173 & causative agent of & disease & vertebrate & 512 & 128 & 640 \\
         UR180 & is finding of disease & disease & body substance & 346 & 87 & 434 \\
         UR211 & biological process involves gene product & gene & function & 650 & 162 & 813 \\
         UR214 & cause of & disease & disease & 459 & 115 & 574 \\
         UR221 & gene mapped to disease & disease & gene & 241 & 61 & 302 \\
         UR254 & may be finding of disease & disease & symptom & 672 & 169 & 841 \\
         UR256 & may be molecular abnormality of disease & disease & genetic aberrant & 244 & 62 & 306 \\
         UR44 & may be prevented by & chemical & disease & 452 & 113 & 566 \\
         UR45 & may be treated by & chemical & disease & 772 & 193 & 965 \\
         UR48 & physiologic effect of & chemical & disease & 700 & 176 & 876 \\
         UR49 & mechanism of action of & chemical & function & 615 & 154 & 769 \\
         UR50 & therapeutic class of & chemical & type & 663 & 166 & 829 \\
         UR588 & process involves gene & gene & disease & 621 & 156 & 777 \\
         UR625 & disease has associated gene & gene & disease & 381 & 96 & 477 \\
        \midrule
        \textbf{Wikidata}\\
        \midrule
        P2175 & medical condition treated & chemical & disease & 621 & 155 & 777 \\
        P2176 & drug used for treatment & disease & chemical & 448 & 112 & 561 \\
        P2293 & Genetic association & gene & disease & 678 & 170 & 849 \\
        P4044 & therapeutic area & chemical & disease & 304 & 76 & 380 \\
        P780 & symptoms & disease & symptom & 289 & 73 & 362 \\
        \midrule
        \textbf{Total} &  &  &  & 19,600 & 4,910 & 24,526 \\
        \bottomrule
    \end{tabular}
    }\vspace{-0.1cm}
    \caption{Detailed statistics of BioLAMA for each relation.
    }
    \label{tab:data_stats}
\end{table*}

\begin{table*}[t]
    \centering
    \resizebox{1.98\columnwidth}{!}{%
    \begin{tabular}{llll}
        \toprule
\textbf{Relation ID} & \textbf{Relation Name} & \textbf{Manual Prompt} \\
\midrule
\textbf{CTD}\\
\midrule
CD1 & therapeutic & [X] prevents diseases such as [Y]. \\
CD2 & marker/mechanism & [X] exposure is associated with significant increases in diseases such as [Y]. \\
CG1 & decreases expression & [X] treatment decreases the levels of [Y] expression. \\
CG17 & increases expression & [X] treatment increases the levels of [Y] expression. \\
CG18 & increases expression & [X] upregulates [Y] protein. \\
CG2 & decreases activity & [X] results in decreased activity of [Y] protein. \\
CG21 & increases phosphorylation & [X] results in increased phosphorylation of [Y] protein. \\
CG4 & increases activity & [X] results in increased activity of [Y] protein. \\
CG6 & decreases expression & [X] treatment decreases the levels of [Y] expression. \\
CG9 & affects binding & [X] binds to [Y] protein. \\
CP1 & decreases & [X] analog results in decreased phenotypes such as [Y]. \\
CP2 & increases & [X] induces phenotypes such as [Y]. \\
CP3 & affects & [X] affects phenotypes such as [Y]. \\
GD1 & marker/mechanism & Gene [X] is associated with diseases such as [Y]. \\
GP1 & association & Gene [X] is associated with pathways such as [Y]. \\
 \midrule
 \textbf{UMLS}\\
\midrule
UR116 & clinically associated with & [X] is clinically associated with [Y]. \\
UR124 & may treat & The most widely used drug for preventing [X] is [Y]. \\
UR148 & due to & [X] induces [Y]. \\
UR173 & causative agent of & [X] is caused by [Y]. \\
UR180 & is finding of disease & [Y] is finding of disease [X]. \\
UR196 & has contraindicated class & [X] and [Y] has a drug-drug interaction. \\
UR211 & biological process involves gene product & [X] involves [Y]. \\
UR214 & cause of & [Y] causes [X]. \\
UR221 & gene mapped to disease & [X] has a genetic association with [Y]. \\
UR254 & may be finding of disease & [X] has symptoms such as [Y]. \\
UR256 & may be molecular abnormality of disease & [Y] has a genetic association with [X]. \\
UR44 & may be prevented by & [X] treats [Y]. \\
UR45 & may be treated by & [X] treats [Y]. \\
UR48 & physiologic effect of & [X] results in [Y]. \\
UR49 & mechanism of action of & [X] has a mechanism of action of [Y]. \\
UR50 & therapeutic class of & [X] is a therapeutic class of [Y]. \\
UR588 & process involves gene & [X] involves [Y] process. \\
UR625 & disease has associated gene & [X] has a genetic association with [Y]. \\
UR97 & contraindicated with disease & [X] has contraindicated drugs such as [Y]. \\
\midrule
 \textbf{Wikidata}\\
\midrule
P2175 & medical condition treated & [X] has effects on diseases such as [Y]. \\
P2176 & drug used for treatment & The standard treatment for patients with [X] is a drug such as [Y]. \\
P2293 & genetic association & Gene [X] has a genetic association with diseases such as [Y]. \\
P4044 & therapeutic area & [X] cures diseases such as [Y]. \\
P780 & symptoms & [X] has symptoms such as [Y]. \\
 \bottomrule
\end{tabular}
    }
    \caption{Manual prompts used in our experiments. Each prompt is created by domain experts.
    \label{tab:manual_prompts}}
\end{table*}
\begin{table*}[t]
    \centering\resizebox{1.95\columnwidth}{!}{%
    \begin{tabular}{llccccc}
    
    \toprule
    
    \textbf{Relation ID}
    & \textbf{Relation Name}
    & \textbf{IE}
    & \multicolumn{2}{c}{\textbf{BioBERT}}
    & \multicolumn{2}{c}{\textbf{Bio-LM}}
    \\

    & & & Manual & Opti. & Manual & Opti. \\
    
    \midrule
    \textbf{CTD} \\
    \midrule
    CD1
    & therapeutic
    & \textbf{14.29}/\textbf{22.33}
    & 3.28/10.16
    & 6.45/16.15
    & 7.20/15.45
    & 7.79/15.51
    \\
    
    CD2
    & marker/mechanism
    & 3.87/6.41
    & 3.28/6.19
    & \textbf{9.81}/\textbf{23.60}
    & 4.64/9.06
    & 6.56/13.44
    \\

    CG1
    & decreases expression
    & 0.15/0.15
    & 0.00/0.00
    & 0.32/0.81
    & 1.16/4.94
    & \textbf{4.59}/\textbf{7.99}
    \\
    
    CG18
    & increases expression
    & \textbf{6.70}/\textbf{19.86}
    & 0.00/0.71
    & 0.00/0.19
    & 0.94/8.58
    & 1.46/7.52
    \\
    
    CG2
    & decreases activity
    & \textbf{8.58}/\textbf{19.49}
    & 0.00/0.00
    & 4.81/8.29
    & 0.67/2.90
    & 4.08/13.41
    \\
    
    CG21
    & increases phosphorylation\
    & 5.04/20.54
    & 0.00/2.71
    & \textbf{14.73}/\textbf{18.14}
    & 0.00/13.95
    & 0.00/15.19
    \\
    
    CG4
    & increases activity
    & 7.83/\textbf{20.83}
    & 0.00/0.00
    & 0.83/2.69
    & 0.00/0.89
    & \textbf{8.42}/20.59
    \\
    
    CG9
    & affects binding
    & \textbf{10.88}/\textbf{23.13}
    & 0.00/0.00
    & 0.23/0.27
    & 0.91/7.94
    & 1.50/5.85
    \\
    
    CP1
    & decreases
    & 0.00/0.00
    & 0.00/0.00
    & \textbf{6.37}/\textbf{19.33}
    & 0.00/1.27
    & 2.35/12.11
    \\
    
    CP2
    & increases
    & 0.00/0.00
    & 0.00/0.14
    & \textbf{10.66}/\textbf{18.38}
    & 0.00/0.81
    & 0.00/0.14
    \\
    
    CP3
    & affects
    & 0.00/0.00
    & 0.00/0.00
    & 0.00/\textbf{1.51}
    & 0.00/0.22
    & 0.00/0.22
    \\
    
    GD1
    & marker/mechanism
    & \textbf{4.17}/8.01
    & 0.33/\textbf{11.64}
    & 0.00/1.67
    & 2.20/8.34
    & 1.43/7.36
    \\
    
    GP1
    & association
    & 1.59/2.04
    & 0.00/17.14
    & \textbf{16.69}/\textbf{31.67}
    & 4.43/21.11
    & 4.00/18.55
    \\
    
    \midrule
    \textbf{UMLS} \\
    \midrule
    UR116
    & clinically associated with
    & 6.35/\textbf{19.28}
    & 0.84/4.07
    & \textbf{7.90}/18.08
    & 2.64/11.02
    & 6.13/14.87
    \\
    
    UR124
    & may treat
    & \textbf{20.52}/\textbf{42.24}
    & 1.03/4.31
    & 1.76/4.10
    & 10.69/25.35
    & 8.79/22.76
    \\
    
    UR173
    & causative agent of
    & 0.31/0.31
    & 1.41/4.84
    & 9.81/\textbf{30.62}
    & 4.53/15.63
    & \textbf{9.84}/27.78
    \\
    
    UR180
    & is finding of disease
    & 0.00/0.23
    & 0.00/0.00
    & 8.57/\textbf{29.72}
    & 0.00/0.00
    & \textbf{9.63}/15.30
    \\
    
    UR211
    & biological process involves gene product
    & 0.00/0.00
    & 0.49/1.85
    & \textbf{12.35}/24.08
    & 0.00/0.25
    & 9.30/\textbf{31.86}
    \\
    
    UR214
    & cause of
    & 1.74/2.44
    & 0.00/1.92
    & 3.83/7.80
    & 1.05/7.32
    & \textbf{3.94}/\textbf{11.88}
    \\
    
    UR221
    & gene mapped to disease
    & 0.00/0.00
    & 0.00/0.00
    & 0.00/0.00
    & 0.00/1.66
    & \textbf{14.44}/\textbf{30.27}
    \\
    
    UR254
    & may be finding of disease
    & 0.00/0.00
    & 10.94/24.26
    & 17.50/37.10
    & \textbf{27.71}/\textbf{38.41}
    & \textbf{27.71}/\textbf{38.41}
    \\
    
    UR256
    & may be molecular abnormality of disease
    & 0.00/0.33
    & 0.00/0.00
    & 0.00/0.00
    & 0.00/0.00
    & \textbf{10.85}/\textbf{19.02}
    \\
    
    UR44
    & may be prevented by
    & 6.89/13.43
    & 1.77/5.65
    & 2.12/7.28
    & 1.24/7.95
    & \textbf{8.83}/\textbf{20.71}
    \\
    
    UR45
    & may be treated by
    & \textbf{17.10}/\textbf{26.22}
    & 1.76/5.80
    & 0.70/4.85
    & 1.76/13.26
    & 8.73/20.39
    \\
    
    UR48
    & physiologic effect of
    & 0.00/0.00
    & 0.00/0.00
    & \textbf{3.06}/\textbf{7.47}
    & 0.00/0.00
    & 1.12/6.03
    \\
    
    UR49
    & mechanism of action of
    & 0.00/0.00
    & 0.00/0.00
    & 0.13/1.14
    & 0.00/0.00
    & \textbf{1.17}/\textbf{3.64}
    \\
    
    UR50
    & therapeutic class of
    & 0.00/0.00
    & 0.12/2.05
    & \textbf{7.17}/14.14
    & 3.50/10.25
    & 6.73/\textbf{21.98}
    \\
    
    UR588
    & process involves gene
    & 0.00/0.00
    & 0.13/1.93
    & \textbf{4.66}/22.47
    & 0.00/1.93
    & 2.60/\textbf{29.73}
    \\
    
    UR625
    & disease has associated gene
    & \textbf{3.56}/7.34
    & 0.00/4.40
    & 1.72/3.61
    & 1.89/\textbf{9.02}
    & 2.26/8.39
    \\
    
    \midrule
    \textbf{Wikidata} \\
    \midrule
    P2175
    & medical condition treated
    & 2.45/7.34
    & 0.64/5.92
    & 3.19/11.04
    & 9.40/21.11
    & \textbf{9.47}/\textbf{24.94}
    \\
    
    P2176
    & drug used for treatment
    & \textbf{22.82}/\textbf{47.24}
    & 1.07/4.10
    & 0.78/9.20
    & 22.46/39.75
    & 20.14/39.57
    \\
    
    P2293
    & genetic association
    & \textbf{9.07}/\textbf{16.61}
    & 0.00/7.77
    & 1.04/4.38
    & 2.24/11.43
    & 2.90/9.21
    \\
    
    P4044
    & therapeutic area
    & 0.26/0.79
    & 4.74/9.21
    & 4.21/8.53
    & \textbf{9.47}/\textbf{19.47}
    & 7.53/18.58
    \\
    
    P780
    & symptoms
    & 0.55/5.80
    & 11.88/29.01
    & 11.82/31.38
    & \textbf{16.30}/\textbf{37.85}
    & 12.98/33.43
    \\
    
    \bottomrule
    
\end{tabular}
}
\caption{Performance on each relation. Acc@1 and Acc@5 are reported. Best performances are in boldface.
\label{tab:result_for_relation}}
\end{table*}
\begin{table*}[t]
    \centering
    \resizebox{2\columnwidth}{!}{%
\begin{tabular}{lll}
\toprule
\textbf{Relation ID} & \textbf{Subject} & \textbf{Top 5 Predictions} \\
\midrule
\multirow{3}{*}{\begin{tabular}[c]{l}CTD - CD1\\ (7.79 / 15.51)\end{tabular}} & \multicolumn{2}{l}{[X] prevents diseases such as [Y].} \\
\cmidrule{2-3}
& Nitric Oxide & \textbf{Hypertension}, Multiple Sclerosis, Cardiac, Pulmonary, Cardiovascular \\
& Triamterene & \textbf{Hypertension}, Epilepsy, Diabetes, Cardiac, Myocardial \\
& SH-6 compound & Epilepsy, Cancer, Liver, Malignant, Inf \\
& quizartinib & Cancer, Liver, Hypertension, Leukemia, Sarcoma \\
\midrule
\multirow{3}{*}{\begin{tabular}[c]{l}CTD - CD2\\ (6.56 / 13.44)\end{tabular}} & \multicolumn{2}{l}{[X] exposure is associated with significant increases in diseases such as [Y].} \\
\cmidrule{2-3}
& Normetanephrine & \textbf{Hypertension}, Cancer, Asthma, Hepatitis, Diabetes \\
& Vitamin K 1 & \textbf{Hypertension}, Hepatitis, Cancer, Diabetes, Anemia \\
& lomefloxacin & Hypertension, Hepatitis, Cancer, Asthma, Diabetes \\
& cefditoren & Hypertension, Hepatitis, Diabetes, Cancer, Asthma \\
\midrule
\multirow{3}{*}{\begin{tabular}[c]{l}UMLS - UR173\\ (9.84 / 27.78)\end{tabular}} & \multicolumn{2}{l}{[X] is caused by [Y] .} \\
\cmidrule{2-3}
& Meningococcal rash & \textbf{Meningococcus}, Streptococcus, Meningococci, Streptococcus pyogenes, Bacteria \\
& Macular syphilide & \textbf{Bacteria}, Virus, T. pallidum, \textbf{Treponema pallidum}, Legionella \\
& Braxy & Bacteria, Virus, Bacterial, Agents, Toxin \\
& Blister with infection & Virus, Adenovirus, Viral, Rotavirus, Enterovirus \\
\midrule
\multirow{3}{*}{\begin{tabular}[c]{l}UMLS - UR211\\ (9.30 / 31.86)\end{tabular}} & \multicolumn{2}{l}{[X] involves [Y] .} \\
\cmidrule{2-3}
& Protein Kinase C & \textbf{Signaling}, Signal, Signal Processing, Apoptosis, Transcription \\
& Guanylate Cyclase & \textbf{Signaling}, Transcription, \textbf{Cell Signaling}, Calcium Signaling, Signal Processing \\
& HLA Complex & Transcription, Immune, Immune Response, Signal Processing, Infection \\
& gephyrin & signaling, Channel Regulation, Receptor Signaling, Signal Processing, \dots \\
\midrule
\multirow{3}{*}{\begin{tabular}[c]{l}UMLS - UR221\\ (14.44 / 30.27)\end{tabular}} & \multicolumn{2}{l}{[X] has a genetic association with [Y] .} \\
\cmidrule{2-3}
& DICER1 syndrome & \textbf{DICER1 gene}, DICER, DICER gene, DICER1, DIC gene \\
& Cervical Wilms Tumor & \textbf{WT1 gene}, WT1, RET gene, PTEN gene, RET \\
& Gangliosidosis GM1 & GM1 gene, GM1, GM gene, gene, GGM1 gene \\
& BALT lymphoma & BCL2 gene, ALT gene, BALT gene, ALK gene, ALK \\
\midrule
\multirow{3}{*}{\begin{tabular}[c]{l}UMLS - UR256\\ (10.85 / 19.02)\end{tabular}} & \multicolumn{2}{l}{[Y] has a genetic association with [X] .} \\
\cmidrule{2-3}
& carcinosarcoma of lung & \textbf{TP53 Gene Inactivation}, TP53 Inactivation, RAS, \textbf{TP53 gene mutation}, EGFR \\
& Liver carcinoma & \textbf{TP53 Gene Inactivation}, TP53 Inactivation, KRAS Inactivation, KIT, RET \\
& Classical Glioblastoma & TP53 Inactivation, TP53 Gene Inactivation, EGFR, RET, MYC Gene Amplification \\
& Intratubular Seminoma & TP53 Inactivation, ERG, TP53 Gene Inactivation, KIT, KIT Inactivation \\
\midrule
\multirow{3}{*}{\begin{tabular}[c]{l}Wikidata - P2175\\ (9.47 / 24.94)\end{tabular}} & \multicolumn{2}{l}{[X] has effects on diseases such as [Y].} \\
\cmidrule{2-3}
& amoxapine & \textbf{depression}, obsessive compulsive disorder, schizoaffective disorder, anxiety, \dots \\
& sofosbuvir & \textbf{chronic hepatitis C}, HIV, AIDS, HCV, hepatitis C virus \\
& duvelisib & AIDS, HIV, cancer, breast cancer, chronic obstructive pulmonary disease \\
& arsenic trioxide & AIDS, diabetes, cancer, tuberculosis, chronic obstructive pulmonary disease \\
\midrule
\multirow{3}{*}{\begin{tabular}[c]{l}Wikidata - P780\\ (12.98 / 33.43)\end{tabular}} & \multicolumn{2}{l}{[X] has symptoms such as [Y].} \\
\cmidrule{2-3}
& legionnaires' disease & \textbf{fever}, pneumonia, fever and cough, cough and fever, \textbf{cough}\\
& Bocavirus infection & \textbf{fever}, conjunctivitis, jaundice, \textbf{diarrhea}, pneumonia\\
& pulmonary tuberculosis & hemoptysis, haemoptysis, dyspnea, cough, chest pain\\
& parenchymatous neurosyphilis & headache, fever, headache and fever, fever and headache, meningitis\\
\bottomrule
\end{tabular}
    }
    \caption{
    Top 5 predictions of Bio-LM (w/ OptiPrompt) given each prompt and different subjects. For each relation, we also report its Acc@1/Acc@5. Correct predictions are in boldface.
    }\vspace{-0.3cm}\label{tab:predictions_appendix}
\end{table*}

\section{Pre-processing of \ours}\label{sec:apdx_prepro}
After applying basic pre-processing steps in~\Cref{sec:biolama}, we aggregate samples with the same subject and relation, which makes each sample contain multiple object answers~(e.g., subj=``COVID-19'', relation=``symptoms of'', obj=\{headache, cough, fever, \dots\}).
We also set the maximun number of triples in each relation as 2,000 while removing relations having less than 500 triples, which are mostly less useful to extract~(e.g., ``affects methylation'' in CTD) or too complicated~(e.g., ``positive therapeutic predictor'' in Wikidata) according to our manual inspection with domain experts.
For the UMLS, out of 974 relations, we select 16 relations that are considered to be the most important by domain experts.
To mitigate the class imbalance problem in object entities, we also undersample highly frequent object entities to be as frequent as the fifth frequent object entity in each relation.

\section{Statistics of \ours }
\label{sec:statistics_of_biolama}

The CTD split has a total of 22,017 samples, the UMLS split a total of 21,164, and the Wikidata split a total of 5,855 samples.
This sums up to a total of 49,036 samples.
\Cref{tab:data_stats} displays the number of samples in each train/dev/test split of each relation.

\section{Manual Prompts}
\label{sec:manual_prompts}

We create multiple manual prompts with the help of domain experts' insight on each relation in \ours~and select the best performing prompts on the development set.
Selected prompts for the relations are listed in \Cref{tab:manual_prompts}.

\section{Implementation Details}
\label{sec:implementation_details}
For confidence-based decoding~\citep{jiang2020x}, we use the open-source code provided by the authors\footnote{\href{https://github.com/jzbjyb/X-FACTR}{https://github.com/jzbjyb/X-FACTR}} and make slight changes for \ours.
We set the beam size to 5 to get the top 5 predictions and the number of masks to 10.
We also set the iteration method to ``None'' as additional iteration did not help to increase the performance.

For OptiPrompt~\citep{zhong2021factual}, we modify the open-source code provided by the authors\footnote{\href{https://github.com/princeton-nlp/OptiPrompt}{https://github.com/princeton-nlp/OptiPrompt}} to allow training over the multi-token objects.
We set the learning rate to 3e-3 and the mini-batch size to 16.
We train OptiPrompt for 10 epochs and select the best checkpoint based on Acc@1 on the development set.
It takes 3 hours to test all samples with manual prompts and 8 hours to train and test with OptiPrompt using 1 Titan X (12GB) GPU.

\section{Result on Each Relation}\label{sec:all_results}
In addition to the averaged performances presented in \Cref{tab:probing_method_experiments}, we present Acc@1 and Acc@5 on each relation in \Cref{tab:result_for_relation}.

\section{More Prediction Examples}\label{sec:predictions_appendix}
We provide more examples on 8 relations where Bio-LM (w/ OptiPrompt) achieves decent top-1 accuracy in \Cref{tab:predictions_appendix}.

\end{document}